\documentclass[sigconf]{acmart}
\usepackage{graphicx} 
\usepackage{algorithm}
\usepackage{algorithmic}
\usepackage{amsmath}

\usepackage{newfloat}
\usepackage{listings}
\usepackage{multirow}
\usepackage{titlesec}
\usepackage{booktabs}


\makeatletter
\renewcommand\@formatdoi[1]{}
\makeatother

\begin{document}
\title{Retrieval Augmented Generation for Domain-specific Question Answering}

\author{Sanat Sharma\textsuperscript{1*}}
\affiliation{%
  \institution{Adobe}
  \country{USA}
}
\author{David Seunghyun Yoon\textsuperscript{1*}}
\affiliation{%
  \institution{Adobe}
  \country{USA}
}
\author{Franck Dernoncourt\textsuperscript{1*}}
\affiliation{%
  \institution{Adobe}
  \country{USA}
}
\author{Dewang Sultania}
\affiliation{%
  \institution{Adobe}
  \country{USA}
}
\author{Karishma Bagga}
\affiliation{%
  \institution{Adobe}
  \country{USA}
}
\author{Mengjiao Zhang}
\affiliation{%
  \institution{Adobe}
  \country{USA}
}
\author{Trung Bui}
\affiliation{%
  \institution{Adobe}
  \country{USA}
}
\author{Varun Kotte}
\affiliation{%
  \institution{Adobe}
  \country{USA}
}

\acmConference[AAAI'24]{Workshop on Scientific Document Understanding,
  February 20--27, 2024, Vancouver, CAN}
  



\renewcommand{\shortauthors}{Sharma, et al.}

\begin{abstract}
Question answering (QA) has become an important application in the advanced development of large language models. General pre-trained large language models for question-answering are not trained to properly understand the knowledge or terminology for a specific domain, such as finance, healthcare, education, and customer service for a product. To better cater to domain-specific understanding, we build an in-house question-answering system for Adobe products. We propose a novel framework to compile a large question-answer database and develop the approach for retrieval-aware finetuning of a Large Language model. We showcase that fine-tuning the retriever leads to major improvements in the final generation. Our overall approach reduces hallucinations during generation while keeping in context the latest retrieval information for contextual grounding.
\end{abstract}

\maketitle
\footnotetext[1]{\textbf{Equal Contribution:} Authors marked with * contributed equally to this work.}

\section{Introduction}

With the advanced natural language processing technologies, large language models (LLMs) have demonstrated impressive capabilities in a wide range of fields, including chatbots~\cite{thoppilan2022lamda}, language translation~\cite{jiao2023chatgpt},
text summarization~\cite{zhang2023benchmarking}, and personalized robot assistance. Question-answering has become an important application for Large Language Models~(LLMs), enhancing the ability of chatbots in diverse domains such as healthcare, finance, education, and customer service~\cite{kaddour2023challenges}. 

However, traditional question-answering systems and general Question-Answering (QA) systems based on LLMs are not trained properly to understand the knowledge and the terminology of the specific field. They will face challenges in effectively retrieving and presenting relevant information, particularly in dynamically updated databases. For example, users' questions about Adobe products are more related to how to use and the characteristics of the products. The question-answering systems outside Adobe products are not trained to properly understand Adobe terminology. Furthermore, the applications and features change frequently and the existing systems often don’t have access to the latest information on Adobe products. Therefore, they cannot provide in-product interactions (e.g., link a menu-item present in the answer to the actual menu
item in the product). Privacy risk is another problem if the data of the product is cloud-based. Furthermore, external solutions have trouble keeping up with the latest changes in Adobe products (e.g., a feature moving from beta to production).

In this work, we develop a new framework to find the answers to the questions that users of Adobe products ask about Adobe products (e.g., "How can I remove all the highlights in a PDF with Acrobat?"). To address these challenges, our approach builds upon a retrieval-based question-answering system. Our smart retrieval
system retrieves the most useful up-to-date relevant content. The retrieval
system is trained in a self-supervised manner based on Adobe data and user behavioral data (clicks). We see that fine-tuning the retriever on the domain-specific documents and click data leads to notable improvement in retrieved results nDCG and quality. We then propose a novel approach for retrieval-aware finetuning of a large language model to obtain the question-answering system for Adobe products. We systematically enhance this system to improve its question-answering performance, ensuring users receive timely and accurate information.

In summary, the contributions of this work can be highlighted as follows:
\begin{itemize}
    \item We propose a novel framework to compile a large question-answer database.
    \item We proposed a full-fledged, advanced, and production-ready retrieval-based question-answering system, in which the retriever is trained based on user user behavioral data (clicks). We showcase how improving the retriever leads to improvement in the final generation.
    \item We develop a novel approach for retrieval-aware finetuning of a Large Language model and use a query augmentation and product disambiguation pipeline to improve retrieval and generation quality on vague queries.
    \item Our overall approach reduces hallucinations during generation while keeping in context the latest retrieval information for contextual grounding.
\end{itemize}

\section{Related Work}

\paragraph{LLM-based question answering systems}

Large language models have been well developed recently and have shown extraordinary performance in many fields and applications. The models with billions of parameters and trained with huge data can assist humans in simple language understanding or generation tasks. To adapt to a specific task or be aligned with human preference, these models usually need to be finetuned~\cite{kaddour2023challenges, brown2020language}. Bakker et al.~\cite{bakker2022fine} finetune the large language models to align the output of the summarization with humans. In the application in our real life, Singhal et al. evaluate PaLM~\cite{chowdhery2022palm} with multiple prompting strategies for medical question answering and comprehension. Using a combination of prompting strategies, Flan-PaLM achieves state-of-the-art accuracy on every dataset in the benchmark MultiMedQA. Yu et al.~\cite{yu2022legal} explore the reason-based prompting mechanism with large language models on the legal question-answering task. Zhou et al.~\cite{zhou2023lima} propose that strong large language models have already acquired knowledge and capabilities. The model only needs a small amount of data for instruction tuning.

\paragraph{Retrieval augmented question answering systems}

Although the large language model can generate fluent or natural text, it might be unfaithful to the source content, i.e., hallucination~\cite{ji2023survey, kaddour2023challenges, maynez2020faithfulness}. Parametric knowledge bias is one of the reasons for hallucination in large language models~\cite{shuster2021retrieval, petroni2019language}. To mitigate the hallucinations, there are some works that focus on incorporating retrieval-augmented architectures into the systems. Lewis et al.~\cite{lewis2020retrieval} proposed RAG, a general-purpose fine-tuning recipe for retrieval-augmented generation. RAG combines pre-trained parametric and non-parametric memory for better language generation. Li et al~\cite{li2023chatdoctor}. proposed a medical chat model, ChatDoctor, finetuned on LLaMA using
medical domain knowledge. The model is finetuned with the medical data from real-world patient-physician conversations and introductions. With the external knowledge base, the ChatDoctor model can retrieve online and offline medical domain knowledge to answer medical questions on up-to-date medical terms and diseases, mitigating the hallucinations and errors in LLMs.

\section{Method}
This section will give more information on our framework. We will first introduce our retrieval system, which retrieves the most useful up-to-date relevant content. Pretrained model confuses Photoshop with Photoshop Express, Premiere Pro with Premiere Rush, etc. We propose query augmentation via product identification to improve overall accuracy and quality of generation. We finetune the large language model based on Adobe data and our trained retriever to understand the user’s question and context and present a useful response.  

\subsection{Framework Overview}

\begin{figure}
  \centering
  \includegraphics[width=\linewidth]{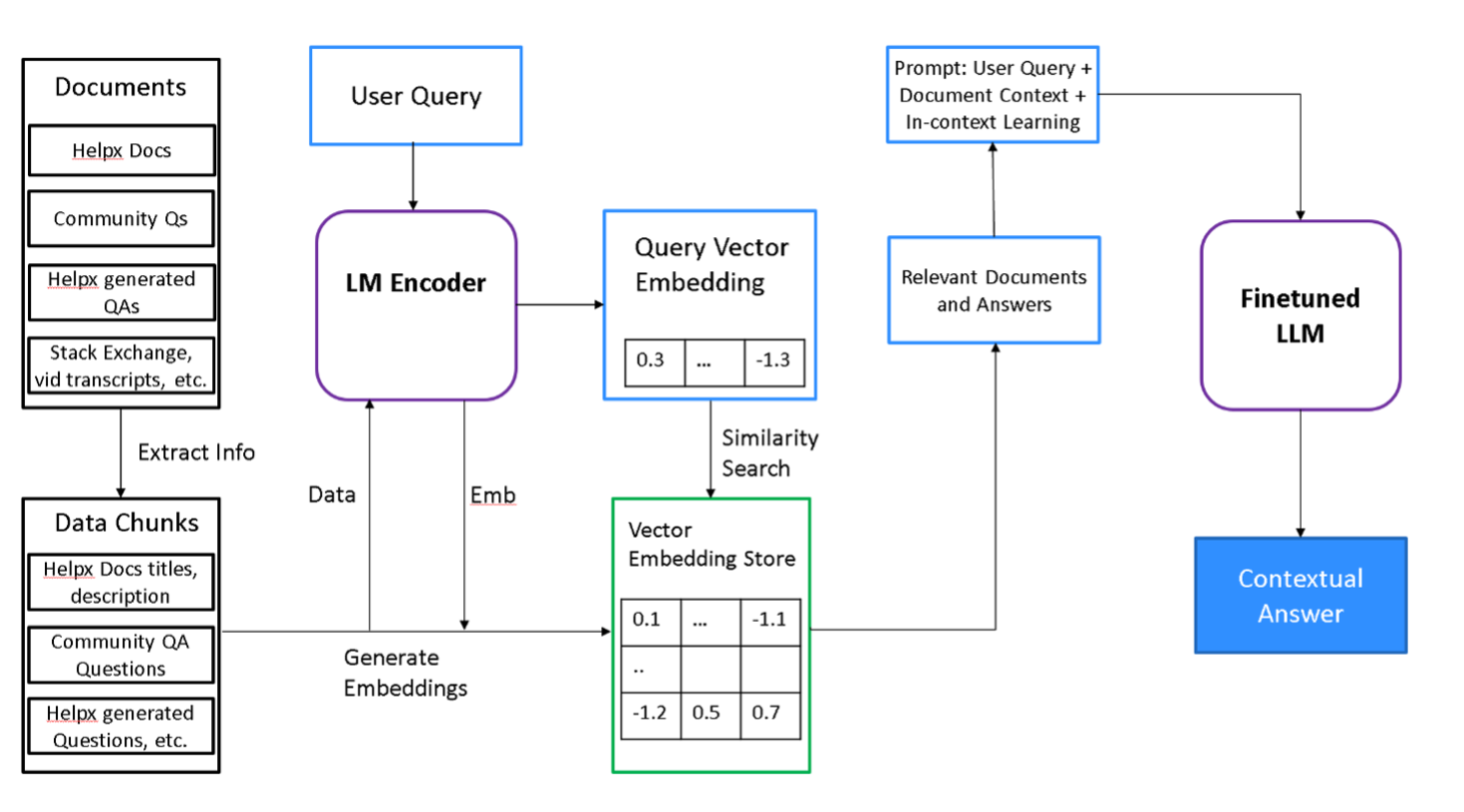}
    \caption{An overview of our proposed framework.}

  \label{fig:overview}
\end{figure}

We use several data sources including Adobe Helpx document titles and descriptions, Adobe Community questions, LLM-generated questions from Helpx documents, and YouTube video transcripts to generate a vector store using our finetuned language model. When a user asks a question, a similarity search is performed to find the closest question matches in our corpus. The most relevant answers along with the user query are passed to our finetuned LLM, which then provides a contextual response to the user. Figure \ref{fig:overview} shows the overview of our proposed framework.

\subsection{Retriever}

\subsubsection{Retriever Training Dataset}
\hfill\

Adobe Helpx (helpx.adobe.com) gets millions of unique visitors every year looking for tutorials and learning content (Adobe Helpx articles, Adobe help videos) related to Adobe products. In order to understand user behavior, we utilize the click logs from user \textsl{queries} $\rightarrow$ \textsl{helpx} and Adobe community content articles to generate our dataset. We use logs from January 2022 – June 2023 for our training and evaluation set. This dataset provides a big and diverse data pool that is critical for learning a good representation of user queries and our retrieval documents. Table~\ref{tab:retriver-data-sample} demonstrates a sample row from the dataset and table~\ref{tab:data_summary} summarizes the overall dataset statistics.

\begin{table*}[]
\caption{A sample row from the retriever training dataset.}
\label{tab:retriver-data-sample}
\begin{tabular}{p{0.13\linewidth}p{0.35\linewidth}p{0.25\linewidth}p{0.15\linewidth}}
\toprule
Query & Document & Doc Input & Log click ratio (relevance) \\ \midrule
Change color of text & \url{https://helpx.adobe.com/indesign/using/editing-text.html} & Learn how to edit text in Adobe   Acrobat & 0.24 \\ \bottomrule
\end{tabular}
\end{table*}

\begin{table*}[]
\caption{The summary of the dataset.}
\label{tab:data_summary}
\begin{tabular}{@{}llll@{}}
\toprule
Dataset Name & Total Rows & Unique Queries & Unique Documents \\
\midrule
Helpx 2022-2023   Click-data & 712792 & 180,799 & 22,576 \\ \bottomrule
\end{tabular}
\end{table*}

For each document, we utilize the Document title and description to represent the crux of the document. Titles are generally 5–20 words long, while the descriptions are generally 20–50 words long, and whole documents can be multiple pages long. Each document in our dataset focuses on specific tasks and hence we found the title and description to be an apt representation of the document instead of passing the full document text.

An example of the title and description is as follows:

\begin{itemize}
    \item Title: How to edit images or objects in a PDF using Adobe Acrobat
    \item Description: Learn how to add, resize, move, or replace images and objects in a PDF using Acrobat. This tutorial showcases multiple ways of editing images in Acrobat using tools such as Flip, Drag, Rotate, Group, Arrange, and more.
\end{itemize}

We also utilize a relevance field that is derived by using the log of the click ratio of the query-document pair, as defined in Equation~\ref{eq:relevance}. $q_i$   represents the specific query. $d_j$ represents the particular document clicked. $\operatorname{clicks}$ represents the number of clicks for the pair and $D_i$ represents the set of documents clicked for $q_i$.

\begin{equation}
    \label{eq:relevance}
    \text { relevance }=\log \left(\frac{\operatorname{clicks}\left(q_i \rightarrow d_j\right)}{\max \left(\operatorname{clicks}\left(q_i \rightarrow D_i\right)\right)}\right)
\end{equation}

This is important since for a query, there may be multiple documents clicked and we wish to pay more attention to higher click pairs. We take the log of the max click ratio since it allows less frequently clicked documents to also be part of the learning process.

\subsubsection{Retriever Model Training}
\hfill\

In order to accurately learn a semantic representation for both queries and documents, we utilize Contrastive Learning. Contrastive Learning is a technique that allows similar samples to be modeled closely in the representation space without having explicit labels. The model is trained in a self-supervised manner, we utilize a single model to represent both the query and the document. This allows us to bring them both in the same representation space. 

We utilize transformers pretrained on sentence similarity tasks (MPnet), due to their proficiency with understanding long texts using attention. We utilize Adam optimizer and mean pooling while training. We also experimented with max pooling, along with first token pooling (similar to cls pooling), and found mean pooling to work best. 

Furthermore, we utilize our relevancy metric (the log of max click ratio in equation~\ref{eq:relevance}) and use it for our weighted cross entropy loss function. This allows us to weight query-document loss based on how relevant the document is for the query. This is important since it adds an inherent ranking signal inside the set of relevant documents.

\begin{figure}
  \centering
  \includegraphics[width=0.5\linewidth]{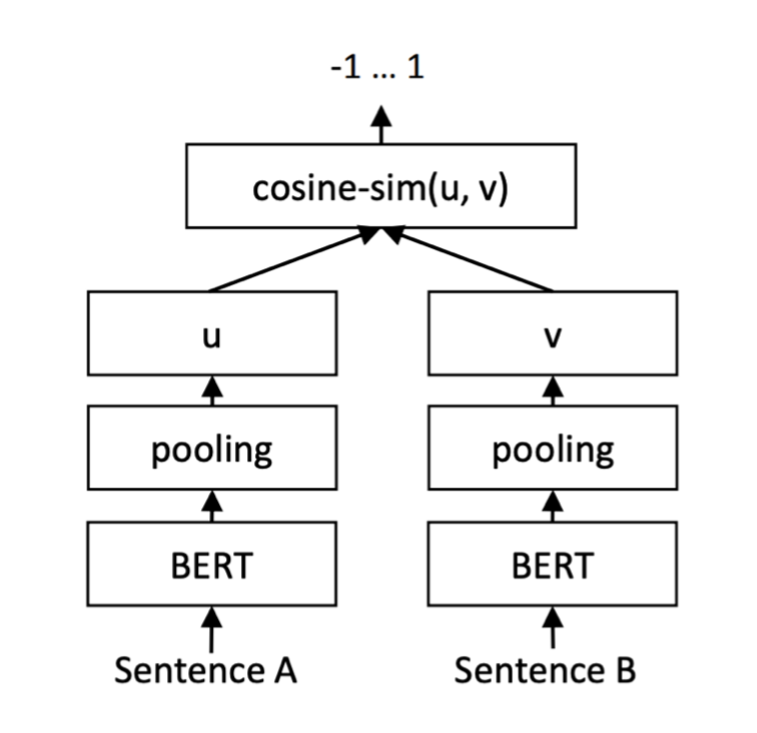}
  \caption{The training processing for the retriever.}
  \label{fig:retriever_train}
\end{figure}

Figure~\ref{fig:retriever_train} showcases our training process. The query and document text are passed together in a training batch to the network and cosine similarities are computed from the generated representations. The mean squared loss is minimized based on the sentence similarity scores. The same weights are shared between the two transformer instances, hence allowing us to learn the representations of both query and document together.

\subsection{Retrieval Index Creation and Database}
\label{sec:Retrieval_Database}

In order to create our retrieval index, we utilize both primary sources (Adobe Helpx documents, Adobe Community questions) as well as derivate sources (generated QA pairs) to provide a rich and diverse retrieval set. We use our finetuned model to generate the representations of all of the sources. We utilize the following sources:

\begin{enumerate}
    \item Primary Sources
    \begin{itemize}
        \item Helpx Documents (title, description) – We take the title and descriptions of helpx documents and embed them.
        \item Community Questions – Adobe has a rich community forum where Adobe users can ask questions related to the product and are helped by Adobe community experts. We utilize this valuable source of data as part of our retrieval index
    \end{itemize}
    \item Derived Datasets
    \begin{itemize}
        \item Helpx Generated QA pairs – In order to generate higher coverage for the Helpx documents and to extract all sub-nuggets of information, we generate multiple QA pairs from each document using a QA generation module powered by a Large Language model.
        \item Adobe Care video QA pairs – Adobe Care is a YouTube channel run by Adobe that provides video tutorials to several highly requested questions as well as providing How-to guides for new features. We extract the transcript from these videos and then utilize our QA generation module to create useful question-answer pairs.
    \end{itemize}
\end{enumerate}

\begin{figure}
  \centering
  \includegraphics[width=\linewidth]{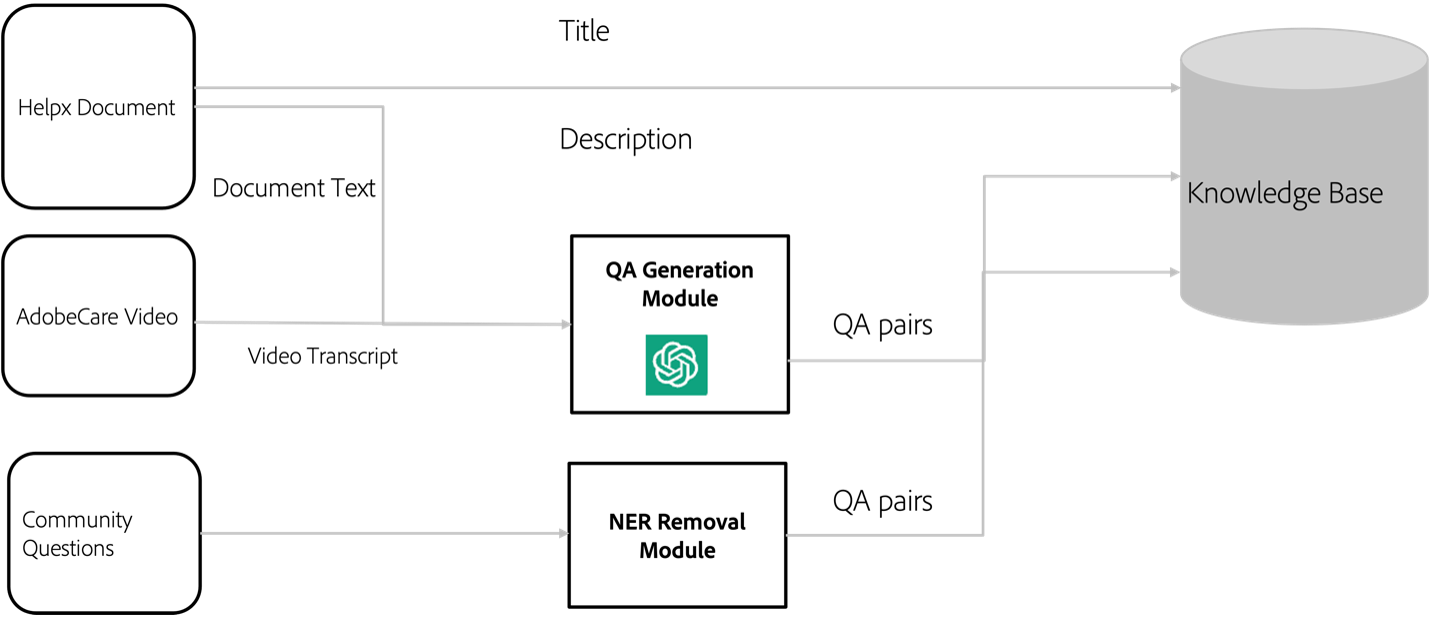}
\caption{An overall architecture for indexing.}
  \label{fig:index_database}
\end{figure}

The overall architecture for indexing is presented in the flowchart of Figure \ref{fig:index_database}.

\subsection{Preprocessing for Building the Database}
In order to curate our derived datasets, we utilize a preprocessing approach to generate question-answer pairs utilizing a QA generation module. In addition, a named entity removal module to provide privacy safeguards for the dataset.

\subsubsection{QA Generation Module}
\hfill\

The QA generation module is designed in order to curate our derived datasets, as discussed in section \ref{sec:Retrieval_Database}. For the Helpx documents and the AdobeCare Video transcripts, we utilize our QA generation module to create the question-answer pairs. The QA generation module is powered by a Large Language Model, with the prompts designed for our task. A sample prompt that we utilized is as follows:

\textit{\textbf{Prompt}: You are an AI assistant that helps create question-answer pairs. You start every question with QUESTION and every answer with ANSWER. Answer in detail.}

We see a significant boost in accuracy in doing few-shot learning compared to zero-shot generation\cite{parnami2022learning}\cite{song2022comprehensive}\cite{brown2020language}. We prepend a couple of sample documents as well as generated QA pairs with each call to our LLM, in order to guide the generation style and quality of the question. In our case, we prefer step-wise responses to questions, as Adobe users tend to prefer discrete steps compared to blocks of text. Table~\ref{tab:QA-example} shows a sample QA pair created by the QA generation module.

\begin{table}[]
\caption{A sample QA pair created by the QA generation module.}
\label{tab:QA-example}
\begin{tabular}{@{}p{\linewidth}@{}}
\toprule
\begin{tabular}[c]{@{}p{\linewidth}@{}}QUESTION: \\ What are the steps to adjust the brightness, contrast, and color in your video clips using the Adjusting Brightness, Contrast, and Color Guided Edit in Adobe Premiere Elements?\end{tabular} \\ \midrule
ANSWER: \\
1. Click Add media to import the video clip you want to enhance. \\
2. Select Guided \textgreater Adjusting Brightness+Contrast \& Color. \\
3. To adjust your video clip, select it. \\
4. Click the Adjust panel to adjust selected settings. \\
5. Click Lighting to adjust the brightness and contrast. \\
6. Click a thumbnail in the grid of the adjustments panel to preview the change in brightness. \\
7. Click More and drag the sliders for more precise adjustment. \\
8. Click Color in the adjustments panel to open the Color section. \\
9. You can adjust the hue, lightness, saturation, and vibrance in the Color tab. \\
10. Click a thumbnail in the grid to preview the change. \\ \bottomrule
\end{tabular}
\end{table}

As part of our indexing pipeline, we automate question-answer pair creation of any new articles and videos in the future, thus allowing us to have a richer retrieval context for all documents in our index.

\subsubsection{Named Entity Removal Module}
\hfill\

In the pipeline of our data-driven ecosystem, the Named Entity Removal Module occupies a critical vertex, ensuring the sanitization of the community-sourced Adobe textual corpus. This module is architected to perform de-identification by removing all classes of Personally Identifiable Information, thereby fortifying the privacy safeguards for the dataset.

The module employs a dual-layered Named Entity Recognition (NER) strategy:

\begin{itemize}
    \item Machine Learning-based NER: At its core, the module integrates a state-of-the-art NER model, \texttt{en\_core\_web\_lg}. This model is pre-trained on a large corpus and employs a deep learning architecture, enabling the efficient categorization of 'PERSON'-type entities with a high degree of accuracy.
    \item Regular Expression-based Custom Sanitization: Augmenting the machine learning-based NER, a custom function, custom\_removal, is deployed to tackle non-standard or pattern-based text segments. This is particularly aimed at recognizing and eliminating textual patterns, email formats, and signature lines. It also includes patterns for phone number sanitization conforming to various international formats.
\end{itemize}
 
The module operates iteratively over the dataset, which is ingested in a JSON-based serialized format. Each entry is processed through the dual-layered NER, and the sanitized text replaces the original content inline, thus ensuring data integrity while enhancing privacy. The serialized output is designed for seamless integration into subsequent machine learning workflows, including our proprietary retrieval and ranking algorithms.
 
This modularity ensures that the Named Entity Removal Module can be seamlessly plugged into various stages of our data processing pipelines, ranging from initial data ingestion to pre-model training sanitization.

\subsection{Query Augmentation via Product Identification}

One of the key challenges we see is related to product disambiguation for vague queries. Since several Adobe products might contain similar features, adding product disambiguation helps improve overall accuracy and quality of generation. Vague queries usually receive vague answers.

We utilize a product-intent extraction model\cite{sharma2024semantic} that maps all input texts to 1 or more Adobe products. We then pass this information to the retriever, allowing for better relevancy in the retrieved documents. 

\subsection{LLM prompting}

Once the retriever has collected one or several passages from the database of question-answer based on the user query, the passages are given as context to an LLM, and the LLM is prompted to answer the user query based on the passages.

The prompt used for the LLM is as follows:

You are an assistant that helps humans use [PRODUCT NAME, e.g. Adobe Premiere Pro]. You will be given a list of question-answer pairs (some pairs might be irrelevant) and a user query. Your goal is to answer the user query using only information from the given question-answer pairs.

List of question-answer pairs: [...]

User query: [...]
Answer: 

In some cases, some pairs in the list of question-answer pairs may be redundant. Therefore, before the list of question-answer pairs is given to the LLM, our system removes duplicates from the list of question-answer pairs. Specifically, we use it to compute the Levenshtein distance as well as the semantic similarity between each question/title and the list of question-answer pairs. If the Levenstein distance is below some threshold or if the semantic similarity is too high, we compute the semantic similarity between their two answers. If the two answers are too semantically similar, we remove one of them based on the source credibility (Helpx > Community > YouTube > LLM-generated QAs) and answer length.

\subsection{LLM Finetuning}

To train the LLM, we use grounded document, $d^{+}$, negative document $d^{-}$, and question-answer pairs, $(q, a)$.  

$y=\text{LLM}_\theta (d^{+},d^{-}, q)$. $y$ is the generated outcome, and it is compared to $a$ to update the LLM. The following shows our template to generate the training samples.


The grounded document, ($d^{+}$), is the original source where $(q, a)$ pair is generated from. To increase the robustness and generalization abilities of the finetuned LLM model, we propose the following approaches:

\begin{itemize}
    \item When the length of the answer $(a)$ is shorter than the threshold (e.g., 90 tokens), we filtered out these examples from the training set to guide the finetuned LLM model, generating an answer with substantial information.
    \item The documents are retrieved from our finetuned retriever model. We use multiple $(d^{+})$ and $(d^{-})$ to increase recall of the retriever components and robustness of the finetuned LLM. Top-k $(d^{+})$s are selected by retrieving documents that are similar to the ground-truth document in the pre-indexed embedding space. 
    \item Negative documents, $(d^{-})$s, are prepared by “random sampling from the database” and by selecting documents that are less similar to the original document (i.e., cosine similarity $< \tau_\text{sim}$) and that are not extremely dis-similar to the original document (i.e., cosine similarity $> \tau_\text{dissim})$.
    \item We also add samples that do not have the grounded documents, $(d^{+})$, but, only have the negative documents, $(d^{-})$. In this case, the answer is set to, “This question cannot be answered at the moment.” This allows the LLM to focus more on the knowledge in a document while training.
\end{itemize}

\section{Experiments}

\subsection{Retriever Evaluation}

For our evaluation, we utilized normalized \-Discounted Cumulative Gain(nDCG) as our metric for two key reasons:
\begin{itemize}
    \item Relevance Grading: nDCG takes into account not only whether a document is relevant but also the degree of relevance. We utilize our relevance numbers from our dataset to assign relevance to each “correct” document (each query might have multiple relevant documents)
    \item Rank Sensitivity: nDCG considers the position of relevant documents in the ranked list. It gives higher scores to relevant documents that appear at the top of the list which is important for us since we can only pass a limited context to our generator. 
\end{itemize}

For our retriever evaluation, we create evaluation datasets per product, verified by human annotators to ensure quality. For each evaluation, we choose queries Adobe users have asked before (sampling on query length and query frequency). We made sure the model has not seen the queries previously in training.

\begin{enumerate}
    \item Acrobat Query-Title Match - Create a 300 query to Helpx document title dataset. We sampled both short and long queries when constructing this dataset. 
    \item Photoshop Query-Title Match - 250 queries to helpx document title dataset.
    \item Lightroom Query-Title Match - 100 queries to helpx document title dataset.
    \item Adobe Express Query-Title Match - 250 queries to helpx document title dataset.
\end{enumerate}

 While computing the nDCG metrics for our finetuned model, we consider all documents in our document set rather than just documents in our evaluation batch to make the task challenging. We compare our retriever to several open and closed sourced options available in industry.

We see significant improvement in retrieved accuracy with our finetuned model, with a smaller model size as shown in table \ref{tab:Retriever_eval}.


\begin{table*}[htbp]
\caption{Retriever performance evaluated on nDCG}
\label{tab:Retriever_eval}
\begin{tabular}{@{}l|lllll @{}}
\toprule
Model & Acrobat (k=4) & Photoshop (k=4) & Lightroom (k=4) & Express (k=4) & Model Size\\ 
\midrule
\textbf{Ours (Finetuned Retriever)} & \textbf{0.7281} & \textbf{0.6921} & \textbf{0.7249} & \textbf{0.8221} & 109m  \\
\midrule
\textbf{BGE-large-en-v1.5 (2023)} & 0.7031 & 0.5680 & 0.6894 & 0.8143 & 335m \\ 
\textbf{UAE-Large-V1 (2023)} & 0.7161 & 0.5601 & 0.6743 & 0.8153 & 335m \\ 
\textbf{DiffCSE (roberta-base-sts, 2022)} & 0.2850 & 0.2571 & 0.3547 & 0.3162 & 124m \\ 
\textbf{SimCSE (sup-roberta-large, 2021)} & 0.5705 & 0.3755 & 0.5786 & 0.6384 & 355m \\ 
\textbf{Clip (ViT-L/14)} & 0.0970 & 0.0349 & 0.3459 & 0.1619 & 427m \\ 
\textbf{OpenClip (ViT-H-14)} & 0.3794 & 0.2867 & 0.4699 & 0.6813 & 986m \\ 
\textbf{T5 (3b)} & 0.1887 & 0.1454 & 0.2959 & 0.1872 & 1.24b \\ 
\textbf{SFR-Embedding-Mistral (2024)} & 0.1623 & 0.1545 & 0.2726 & 0.1377 & 7b \\ 
\textbf{mxbai-embed-large-v1 (2024)} & 0.7150 & 0.5526 & 0.6783 & 0.8175 & 335m \\ \bottomrule
\end{tabular}
\end{table*}

    

We see significant improvement in retrieval quality for both head and tail queries in our evaluation. Two major improvements include:

\begin{enumerate}
    \item Domain Understanding – One of the key attributes gained is the ability to understand product names and place importance on them. Pretrained model confuses Photoshop with Photoshop Express, Premiere Pro with Premiere Rush, etc.
    \item Ability to comprehend shorter sentences. Our user behavior shows that users often type sentences < 15 words to get the data they need. Our model outperforms the baseline on shorter sentences.
\end{enumerate}

The total number of items indexed in the retrieval index database is 121,547. The data statistics are listed in table~\ref{tab:database}.

\begin{table}[]
\caption{Summary of the retrieval index database.}
\label{tab:database}
\begin{tabular}{@{}lll @{}}
\toprule
Dataset Sources & Num Rows & Dataset \% \\ \midrule
Helpx Articles & 64,959 & 53.4\% \\
Community Questions & 15,148 & 12.5\% \\
Generated Helpx QA & 40,909 & 33.7\% \\
Generated AdobeCare Video QA & 531 & 0.4\% \\ \bottomrule
\end{tabular}
\end{table}

\subsection{Query Augmentation via Product Identification}
To address the ambiguous queries across various Adobe products and enhance the overall accuracy and quality of generated responses, we utilize a product-intent extraction model that maps all input texts to effectively maps all input texts to one or more relevant Adobe products. This extracted product information is passed into the retriever, thereby enhancing the relevance of retrieved documents.

Table~\ref{tab:query_augmentation} shows the generated answers for retriever that with and without product identification. Before using the product intent model, which means without passing the product identification to the retriever, we get general responses since there are pdf templates in multiple products such as Illustrator, Indesign and Acrobat. By providing product disambiguation, we focus on Adobe Acrobat which is the most common product used with PDFs.

\begin{table}
\centering
\caption{The generated response with and without product identification for the retriever.}
\label{tab:query_augmentation}
\begin{tabular}{lp{0.75\linewidth}} 
\toprule
\multirow{2}{*}{without identifier} & how do I create a blank PDF \\ 
\cmidrule(lr){2-2}
 & To create a blank PDF, select File > New From Template, open the Blank Templates folder, select a template, and select New. Illustrator creates a new document based on the selected blank template. \\ 
\midrule
\multirow{7}{*}{with identifier} & how to create a blank PDF \\ 
\cmidrule(lr){2-2}
 & To create a blank PDF, you can use Adobe Acrobat. Here are the steps: \\
 & 1. Open Acrobat and choose “Tools” > “Create PDF”. \\
 & 2. Click on “Blank Page”. \\
 & 3. Select the size of your blank page. \\
 & 4. Click on “Create” and the new blank PDF will open in Acrobat for editing or saving. \\
 & I hope this helps! Let me know if you have any further questions. \\
\bottomrule
\end{tabular}
\end{table}

\subsection{Quantitative Evaluation}

In addition to doing component specific evaluation, we also evaluate the full system both quantitatively and qualitatively. For our quantitative evaluation, we focus on Adobe Acrobat help questions. We utilize 137 user-asked questions and utilize Acrobat product-experts to generate gold responses for each questions. The questions include short common questions such as "\textit{How to remove password from the PDF}", as well as longer questions such as "\textit{I have a large size pdf which I a unbale to share. how do I compress it and share this compressed pdf with others}".

For the evaluation questions, we generate responses from all candidates and then utilize GPT4\cite{openai2023gpt4} to evaluate the similarity of the candidate-generated answer with the gold human-annotated answer on a scale of 0-1. For GPT-4, we set 'n = 20, temperature = 1, top p = 1' to sample 20 times to estimate the token probabilities\cite{liu2023geval}. Finally, we take the mean of all scores for a row. The relevance scores are showcased in table \ref{tab:quant_eval}.

We see that improving the retriever does affect overall quality of the generations as well. For in-domain tasks, we recommend finetuning the retriever to enhance overall generation performance. We also see poor performance of the finetuned generator on the test set. We ascribe this due to a gap between LLM-generated qa pairs (trainset) and human-generated ground-truth writing style.

\begin{table*}[]
    \centering
    \caption{Generation Relevance on Acrobat test set.}
    \begin{tabular}{{@{}ll@{}}}
        Model & Relevance Score (0-1) \\
       \toprule
       Finetuned Retriever + Finetuned Generator & 0.5150 \\
       Finetuned Retriever + GPT 4 & \textbf{0.7242}\\
       \midrule
        GPT4 & 0.1705\\
       SFR-Embedding-Mistral + GPT4 & 0.7022\\
       sentence-transformers/multi-qa-MiniLM-L6-cos-v1 + GPT4 & 0.5625 \\
       mxbai-embed-large-v1 + GPT4 & 0.6617 \\
       \bottomrule
    \end{tabular}
    \label{tab:quant_eval}
\end{table*}

\subsection{Qualitative Evaluation - Comparison of the generated answers}

In this section, we’ll review a few examples comparing the output of our system with OpenAI’s ChatGPT\footnote{ (\url{https://chat.openai.com}) }.

\paragraph{Example 1}
\begin{itemize}
    \item \textit{Question:} What is firefly?
    \item \textit{Answer from ChatGPT:} Fireflies are a type of bioluminescent insect known for their ability to produce light. They belong to the family Lampyridae, which is part of the order Coleoptera, the beetles. Fireflies are also commonly referred to as lightning bugs in some regions, particularly in North America. […]
\end{itemize}

OpenAI’s ChatGPT does not answer about Adobe Firefly. Even if the question is expressed more specifically, ChatGPT does not know the answer. By contrast, the answer to our system is more accurate.

\begin{itemize}
    \item \textit{Question:} 
    What is Adobe Firefly?
    \item \textit{Answer from ChatGPT:} 
    
    As of my last knowledge update in September 2021, there is no widely recognized software or product called "Adobe Firefly" from Adobe Inc. Adobe is known for its extensive range of software products, including Adobe Photoshop, Adobe Illustrator, Adobe Premiere Pro, Adobe Acrobat, and many others, which are commonly used for design, video editing, document management, and various creative and productivity tasks. […]

\end{itemize}

\begin{itemize}
    \item \textit{Answer from our system:} 
    
    Firefly is a family of creative generative AI models coming to Adobe products, with the ability to use everyday language to generate extraordinary new content. It offers unlimited creative options for digital imaging and photography, illustration, artwork, graphic design, video, marketing and social media, and 3D modeling. It can generate custom vectors, brushes, and textures from just a few words or even a sketch, and can change the mood, atmosphere, or even the weather in a video. It also offers generative image compositing and the ability to turn simple 3D compositions into photorealistic images.

\end{itemize}

\paragraph{Example 2}
Another key area where retrieval is important is when pricing information is needed. Since pricing and offers change over time, retrieving the latest information is key.
ChatGPT provides a general answer to the question whereas our solution is able to provide the latest numbers.

\begin{itemize}
    \item Question: How much does Premiere Pro cost for students?
    \item \textit{Answer from ChatGPT:} 
    
    As of my last knowledge update in September 2021, Adobe offers a special pricing plan for students and teachers called Adobe Creative Cloud for Education. The cost for Adobe Premiere Pro, as part of this plan, varies depending on the subscription you choose and your location. Here are some general pricing guidelines for students:
    
    Adobe Creative Cloud All Apps Plan: This plan includes access to all Adobe Creative Cloud applications, including Premiere Pro. The student/teacher pricing typically ranged from around \$19.99 to \$29.99 per month, depending on the specific promotional offers available at the time. Adobe occasionally offers discounts or promotions for students, so it's a good idea to check their website for the most up-to-date pricing.
    
    Premiere Pro Single App Plan: If you only need access to Premiere Pro, Adobe also offers a single app plan. As of my last update, this plan for students and teachers was priced at around \$20.99 per month.
    \item \textit{Answer from our system:} 
    
    Premiere Pro is available for students as a part of Adobe’s Creative Cloud All Apps plan, which includes 20+ apps like Adobe Photoshop, Acrobat, and Illustrator. The regular price for this plan is US\$54.99/month, but eligible students can get it at a discounted price of US\$19.99/month for the first year and US\$29.99/month after that, which is over 60\% off. This plan also includes 100GB of cloud storage and access to hundreds of thousands of royalty-free assets via Adobe Stock for commercial or personal use. Students can also try Premiere Pro for free with a 7-day trial of the Creative Cloud All Apps plan. For more information, please see Adobe’s terms and conditions or compare plans and pricing.
\end{itemize}


Note that answer quality aside, our system has many other upsides over using ChatGPT: privacy, cost, updated to incorporate the latest information about Adobe products, our system can have hyperlinks in the answers or even links to panels/windows/tools/etc. within the product, and we can gather user feedback to improve our answers over time. This makes an in-house question-answering solution much preferable to just using ChatGPT.

\section{Conclusion}
In this paper, we introduce a novel Question-Answering system for Adobe products, including the retriever and a generator. We use our Adobe documents, such as Helpx documents as the retriever training dataset. Furthermore, we derive a metric for relevance and use it for the weighted cross entropy loss function when training the retriever. We build our retrieval database with multiple sources of data. We remove the personal information and generate QA pairs for the database in the preprocessing. To address the challenge of disambiguation for vague queries, we add the product-intent information to the retriever. We train our QA system using grounded documents, negative documents, and question-answer pairs. We compare our QA system with the openAI ChatGPT on some questions related to Adobe products. The experiments show that while ChatGPT fails to answer the questions correctly or generates some useless information for the user, our system can generate up-to-date, concise, and correct answers.


\bibliographystyle{plain}
\bibliography{sample-ceur}




\end{document}